# A Review on Explainability in Multimodal Deep Neural Nets

Gargi Joshi[1], Rahee Walambe[2,*] and Ketan Kotecha[2,*]

[1] Research Scholar, Symbiosis Institute of Technology (SIT), Symbiosis International (Deemed University), Pune, India.
[2] Symbiosis Centre for Applied Artificial Intelligence, Symbiosis International (Deemed University), Pune, India

Corresponding author: Ketan Kotecha (email: director@sitpune.edu.in) and Rahee Walambe (e-mail: rahee.walambe@scaai.siu.edu.in)

"This work was supported by the MHRD SPARC Grant No P104 of the Government of India."

**ABSTRACT** Artificial Intelligence techniques powered by deep neural nets have achieved much success in several application domains, most significantly and notably in the Computer Vision applications and Natural Language Processing tasks. Surpassing human-level performance propelled the research in the applications where different modalities amongst language, vision, sensory, text play an important role in accurate predictions and identification. Several multimodal fusion methods employing deep learning models are proposed in the literature. Despite their outstanding performance, the complex, opaque and black-box nature of the deep neural nets limits their social acceptance and usability. This has given rise to the quest for model interpretability and explainability, more so in the complex tasks involving multimodal AI methods. This paper extensively reviews the present literature to present a comprehensive survey and commentary on the explainability in multimodal deep neural nets, especially for the vision and language tasks. Several topics on multimodal AI and its applications for generic domains have been covered in this paper, including the significance, datasets, fundamental building blocks of the methods and techniques, challenges, applications, and future trends in this domain.

## I. INTRODUCTION

Remarkable improvements of deep neural nets in the independent tasks based on several modalities such as vision, sensor data, textual data, and language have given rise to many new trends and applications in the integrated space of deep multimodal learning [1]. Deep neural nets[2] have proved exceptionally effective for several single modality [3] or multimodality tasks [4]. However, the complex hidden layers processing in the deep neural nets makes them difficult to interpret, opaque, and black-box models with little or no understanding of their internal states and decision-making process [5]. Gaining meaningful knowledge and understanding of how and why the model arrived at a particular decision or outcome is crucial in model explainability, making it one of the important evaluation metrics [6]. The lack of understanding of the underlying process questions the model's credibility and transparency, impacting its social acceptability and usability [7], [8]. In a multimodal environment with diverse modalities having varied scales and representations, the extraction of information from various heterogeneous sources is essential for the integration and fusion of these modalities. Several applications and tasks for multimodal AI are proposed in the literature combining various modalities. However, the most prevalent applications are found in vision and language tasks. We have considered the image and video modalities in the vision tasks and text and audio modalities in language tasks in this work.

Multimodality extracts and combines vital information from the respective modality source and solves a given problem with richer representation and performance than the individual modalities [9]. The complementary nature of modalities is explored in [10] to complement the missing data or noise in modalities The inter, intra and cross-modal interactions, correlations, and relationships between multiple modalities with high mutual information are explored to have improved predictions and performance. An optimal fusion scheme would combine the modalities and ensure that the resultant model reflects the salient features of input modalities to generate a rich, joint representation for a downstream task [11]. Vision and language models exploit cues in the question and biases in the data distribution with minimum dependence on visual content leading the model to answer a task. However, due to the non-consideration of vision info, the model does answer the questions with high confidence even though they are wrong with intrinsic weaknesses. In Visual Questioning Answering to diagnose the model's performance, the inverse Visual Question Answering task is proposed in a multimodal setting [12]. Answering questions without proper grounding and pointing to evidence hampers performance with multiple modalities [13]. The commonly used method for interpreting the multimodal setting results is Grad CAM's attention maps [14]. Still, they do not identify whether the model looks at the correct region to answer the questions. Attention maps do not guarantee explanatory power [15] as the models and human perception does not concentrate on the same areas while providing the output. Lack of alignment in the image-text pairs emphasizes that mere accuracy is not enough if it is not accompanied by a valid justification, i.e., to be right for the right reason [16]. Explainability efforts are also carried out by generating visual [17] and textual explanations[18], but they can result in multiple explanations with varied evaluation metrics. Modular approaches make the system interpretable by design[19], [20] but are majorly tested on synthetic datasets such as CLEVR [21]. To exploit the complementary assistive and explanatory information and improved predictive power from multiple modalities, a deep understanding of the model's working, predictions, and flaw detection are of utmost importance by using the respective modalities [22].

This growing critical need and importance have led to several reviews emphasizing the multifaceted explainability topic in black-box models such as deep neural nets [23]. The existing literature has a number of surveys on explainable AI such as [24][25], [5],[8],[26][27] and on explainable deep learning [28],[29]. In [25], Gilpin et al. suggested a taxonomy for explaining the explanations by classifying the XAI methods based on- processing, representation, and the type of explanation produced with their evaluation based on completeness to model and substitute tasks. In [29], a survey of different methods on understanding and interpreting deep neural nets are discussed. In [30], a survey of explainable deep learning applications for medical imaging tasks from the perceptive of deep learning researcher, developer, and end-user is presented. [31] presents the categorization of different explainability methods applicable to the medical and healthcare sector. In [32], a survey on interpretability methods in machine learning from a causal perspective is presented. Graph neural nets show a key role in explainability from a causal perspective. Constructing a multimodal feature representation space spanning diverse modalities from a causal and counterfactual perspective in the medical domain is detailed [33]. [34] presents an overview of the Explainable AI (XAI) approaches for the Natural Language Processing domain. Multiple deep learning and embedding approaches are reported for NLP tasks such as text mining [35],[36], sentiment analysis [37],[38],sarcasm identification [39],[40], information extraction [41],[42],[43]. Visual analytics [44] plays a vital role in understanding the deep neural net models using different techniques such as node-link diagrams, dimensionality reduction, line charts, temporal metrics, and instance-based analysis; graphical methods analyze model parameters, individual computational neurons, and activation units. Visual representations, interactions, attribution, and feature visualization techniques provide interpretability, scalability, bias, and adversarial attack detection covering different visualization techniques. Existing surveys in the field present either generalized or specific perceptive to XAI methods, techniques, and applications from the unimodal context. In contrast, we focus on the applicability of explainable AI in the multimodal setting involving diverse heterogeneous multiple modalities that have not been previously addressed and are key to this work. This review encounters explainability in multimodal tasks with a specific focus on the vision and language tasks, where interpretability in the model is established using disentangled representations, multimodal explanations, and counterfactuals techniques [25].

### A. ORGANIZATION OF THE SURVEY

This paper provides an introduction to explainability in deep neural nets with a specific focus on the explainability in the multimodal setting. We consider and discuss deep multimodal learning for different vision and language tasks and typical challenges in the multimodal environment. The methods and techniques of multimodal data fusion and integration are discussed in Section 2. Different types of explainability techniques in unimodal and multimodal settings are covered. The taxonomy of varying XAI techniques is presented in Section 3. The significance of explainability in multimodal networks with introspection and justification systems is discussed in Section 4. Different multimodal explanation techniques such as attention-based approaches, counterfactual explanations, interactive approaches, graph-based approaches, attribute-based techniques are clearly and distinctly classified and discussed in Section 5. Explanation evaluation methods based on the human mental model and automated processes are reviewed in Section 6. Different explainability requirements to satisfy the needs and expectations of different stakeholders are analyzed in Section7. Widely used datasets for explainability in multimodal networks and applications are listed in Section 8. Topics aligned with multimodal explainability setting such as multimodal bias and fairness, adversarial attacks that enhance robustness and interpretability are discussed in subsequent sections in Section 9 and Section 10. Finally, we conclude the survey outcomes as observations and recommendations that showcase the gaps and findings with the further scope of improvement and persisting challenges, future research trends, and directions to pave a roadmap for further research in this active domain in Section 11 and Section 12 respectively.

### II. MULTIMODAL MACHINE LEARNING

The McGurk effect[45] highlighted that the audio and visual information is merged into a unified, integrated perception that led to audio-visual speech recognition (AVSR) systems, giving rise to multimodal multisensory interfaces and multimodal information retrieval systems. Large scale datasets, faster GPUs, visual and language features are key enablers of multimodal machine learning research in the deep learning era [46].

Human perception is multimodal. Humans have the inherent cognitive ability to relate and process information from multiple heterogeneous modality sources at a single instance through the senses. We perceive and tackle things in a multimodal way. Modality is the form in which information is stored or represented and is conveyed through a media [46]. With the recent technological advances, the data come from a diverse number of sources. For example, data on social media sites is heterogeneous, high dimensional, complex, and is represented by multiple modalities such as image, text, audio, and video in the verbal, vocal, and visual form [47]. These diverse modalities differ in their scales, representation format, varied predictive power, weights, and contributions towards the final task [9]. Optimal data fusion schemes such as early [11], late [48], and hybrid fusion [49] schemes are developed to fuse the modalities at data, feature, decision, and intermediate mixed levels [50]. Deep neural nets [51], kernel-based methods [52], and graphical models [47, 48] are employed for analysis and handling such data depending on the downstream task [46]. Individual modalities are mapped onto a common shared representation vector space either through joint or coordinated representation [55].

Multimodal setting leverages improved predictive power compared to its unimodal counterparts due to involvement and knowledge extraction from multiple modalities. This achieves improved results and richer representation with task-relevant features, reducing data size compared to single modality representations [4]. The interplay between multiple heterogeneous and high dimensional diverse modality sources with diverse representation formats makes explainability a key concern for multimodal data. This leads to deriving comprehensive global insights about the model's design, working principle, decision-making process, flaw detection, and handling the bias and fairness issues [56]. Multimodal data provide complementary, additive, combined, and comprehensive information exploring the inter, intra, shared, and cross-modality associations and correlations between different modalities [57].

Several applications are reported in literature where multiple modalities are leveraged. Table I enlist such applications in various generic domains. Table II refers to a number of multimodal applications applied to critical domains namely healthcare, autonomous robots, finance as there is a growing use of AI in these domains leading to a tremendous need for explainability for social acceptance and usability.

TABLE I
MULTIMODAL APPLICATIONS APPLIED IN VARIOUS GENERIC DOMAINS

| Task | Reference | Modality | Remarks |
|---|---|---|---|
| Affect Recognition | [58] | text, audio-video, image, physiological signals conversations | The field in which emotions, personality, and sentiments are identified and studied with multiple modalities |
| Media Description | [49] | image, audio, video, text | Provides fusion strategies for multimedia analysis |
| Image Captioning | [59] | image, text | Deep learning-based methods for generating image captions for control systems and IoT devices |
| Video captioning | [60] | audio, video, text | Survey and evaluation of the deep learning-based image and video captioning |
| Visual Question Answering | [61] | image, text | Discussion on Visual Question Answering in which a natural language question is answered based on its image content. Methods and datasets for VQA are discussed in detail |
| Multimedia information retrieval | [62] | image, audio, video, text | Discussion on the extraction of semantic information from multimedia data sources |
| Visual Common Sense reasoning | [63] | image, text | Provides answer and rationale of a visual question answering system |
| Vision Language Navigation | [64] | image, text | Discussion on decision making based on a reinforcement learning based model for the visual navigation system |
| Biometrics | [65] | visual | presents a survey on multimodal biometric authentication and template protection |
| Human activity recognition | [66], [67] | Accelerometer and other sensors | Predicting the movement of a person based on the sensor data |
| Transportation Systems | [68] | various modalities | Data fusion techniques applied to various transportation systems |
| Personality trait analysis | [69] | Text, image, video | Automatic analysis of job interview screening decisions analysis of patterns in acting, feeling, and thinking process |

TABLE II
MULTIMODAL APPLICATIONS IN CRITICAL DOMAINS

| Domain | Reference | Modality | Remarks |
|---|---|---|---|
| Autonomous driving | [70] | RGB, RGBD, LiDAR | Modular and end-to-end autonomous driving with several sensor modalities is discussed. |
| | [71] | Camera, LiDAR, Radar | Paper provides multimodal learning perceptive to advanced driving systems. |
| | [72] [73] | Visual, text, Sensory | Interpretability in the system is achieved through the introduction of a visual attention mechanism with improved performance |
| | [74] | Camera, LiDAR, Radar, IMU, GNSS | Task-related to perception, localization planning, and decision making are reviewed with the tools and datasets. |
| | [75] | Camera, LiDAR, Radar | Tasks related to deep object detection and segmentation with associated fusion challenges are detailed and addressed. |
| | [76] | Visual, Sensory | Explainability for vision-based autonomous driving systems is detailed out. |
| | [77] | Visual, Sensory | The visualBackProp method is proposed for visualizing the necessary image pixels for interpreting the predictions of the CNN. |
| | [78] | Visual, Sensory | Paper uses a causal approach to visual attention mechanism. |
| | [79] | text | Paper provides textual based explanations for a self-driving car |
| Healthcare | [80] | X-Ray, CT, Text (Reports), USG, MRI, ECG | The paper focuses on the causal aspect in the multimodal healthcare domain with counterfactual explanations. |
| | [81] | | Explainability methods for the critical domain of healthcare are discussed. |
| | [82] | | The paper focuses on studying and mapping brain functionalities and explainability in the neuroscience domain. |
| | [31] | | A detailed survey on XAI in the medical domain is presented. |
| | [83] | | Paper surveys and evaluates explanation from clinician trust wise in the medical domain. |
| | [84] | | Paper review XAI in the medical domain from multimodal and multicenter points of view. |
| | [85] | | The paper highlights the advancement and challenges of XAI in computer-aided diagnosis. |
| | [86] [87] [30] | | Healthcare assistive technologies |
| | [88] | | Medical image analysis Genetics and biomedicine for the integration of omics data. |
| Finance | [89] | Tabular, Temporal | Paper generates user-friendly explanations for loan denials using GAN |
| | [90] | | Paper carried out multimodal fusion techniques for forecasting international stock markets. |

However, due to persistent heterogeneity in multimodal data, it remains challenging to develop new and efficient analytical methodologies. It is a complex task to deal with the multimodal data to explore their comprehensive benefits that have attracted rich attention in the research domain in recent years [91]. Hence there is an urgent need to understand the approaches, methods, and techniques for multimodal data fusion and build an integrative framework for developing tools and applications in various disciplines [92]. Interpretability, explainability, and contextual cognitive reasoning assist in understanding multimodal data in a better way [93]. With the development of new benchmarks, we can identify and handle the flaws in model evaluation metrics, dataset bias, robustness, and spurious correlations [94] in the multimodal data.

## A. TYPICAL CHALLENGES IN MULTIMODAL SETTING

Multimodal AI is inherently complex. Following broad-scale challenges are identified [40].
1) *Representation* is the method or format in which modalities are represented that extracts the complementary or redundant information between multiple modalities. Due to the heterogeneous nature of multimodal data, its representation is very important and, at the same time, challenging too. For example- the sound is a signal, and the image is a 3D representation with varied scales and dimensions to represent. How to bring them into the same common representation space is an essential aspect of implementation.
2) *Translation* refers to the process of explaining how-to transform or convert data from one modality to another when heterogeneous. The relationship between modalities is often subjective. For example, translating a video to its corresponding text description.
3) *Alignment* is the mapping of the direct correlations between sub-elements from two or more different modalities. For example, we may want to align the streaming video and its subtitles. To overcome this challenge, we measure the similarity between other modalities and deal with possible long-range dependencies and ambiguities. Alignment and representation can be considered as overlapping tasks with a marginal difference.
4) *Fusion* is the method of integrating or fusing information from two or more modalities to perform a prediction. For example- for audio-visual speech recognition, the lip motion's visual description is combined with the speech signal to predict spoken words. The information coming from different modalities may have varying predictive power, importance, contribution, and noise topology, possibly missing data in at least one of the modalities.
5) *Co-learning* is the process of sharing knowledge and information between multiple modalities. It is primarily applied when one modality is a rich source of information, whereas the other is a poor information source. The knowledge between modalities can be exchanged and shared to enrich the modeling process, such as in Transfer Learning.

Multimodal data fusion is often prone to overfitting as we have data from various sources and variable learning rates that generalize differently. The overfitting problem can be addressed by implementing a dynamic and adaptive fusion scheme and a regularization technique such as gradient blending that computes the optimal blend of modalities based on their overfitting behavior [95]. Explainability in the multimodal setting is vital to have a comprehensive global view of data and address the associated challenges [96]. Table III describes modality representations, architectural requirements, and pre-trained models in multimodal vision and language paradigm.

TABLE III
REPRESENTATION AND PRETRAINING OF VISION AND LANGUAGE MODALITIES

| Modalities | Approaches | Pretraining |
|---|---|---|
| Image | Region-based CNN (R-CNN) for learning local features in images [97] | Transformer BERT and its variations |
| Text | Unidirectional long short term memory(LSTM)[104], Bidirectional LSTM (BiLSTM), Unidirectional Gated Recurrent Units (GRU)[105],bidirectional GRU (BiGRU) | BERT Transformer (B2T2) [98], Unicode-VL [99], VL-BERT[100], UNITER [101], ViLBert [102], LXMERT [103] |

## B. MULTIMODAL DATA FUSION TECHNIQUES

Depending on the level at which the fusion of input modalities occurs in the network, the multimodal data fusion mechanisms are classified as early fusion (data or feature level fusion, late fusion (decision level fusion), and hybrid (intermediate or joint) fusion. The fusion mechanisms are highly data, task, and application-specific; hence, the appropriate and optimal fusion mechanism is critical. The data fusion methods are broadly classified into early, late, and hybrid fusion approaches.

1) EARLY FUSION

Early fusion is a traditional way of integrating data before its analysis. It has two methods. The first method is fusing data by removing the correlations. The second method is to combine data at its lower-dimensional latent subspace [106]. Statistical solutions such as Principal Component Analysis [107], Independent Component Analysis [108], and Canonical Component Analysis [109] are proposed for fusion by reducing the data dimensions. Early fusion is applied and performed on unprocessed raw data. Features are extracted before fusion for modalities with variable sampling rates to avoid complexity. Syncing of data sources is also difficult when they are either in discrete and continuous forms. So, converting data sources into a fixed representation is too difficult and time-consuming with early fusion techniques [110]. Early data fusion is assumed to be conditional dependent, but there is a high correlation among multiple modalities; for example, MRI is associated with depth. Modalities are associated at a higher level of abstraction [110]. The outcomes of individual modality are expected to be processed irrespective of each other. Features are fused using simple concatenation [111], pooling, and gated units [112]. In reality, modalities have different dimensions for fusion. The major drawback of using early fusion is that- it removes a large amount of data from the modalities before fusing it. The method fails in the synchronization of time stamps. This problem can be solved if we collect the data at similar sampling rates. Different solutions to overcome this problem are proposed, combining sequential and discrete events with continuous data through training, pooling, and convolution [113]. An early fusion approach is schematically represented in Fig 1.

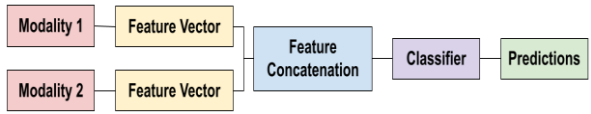

**FIGURE 1.** An early fusion approach

2) LATE FUSION

Late fusion uses individual modality sources for fusion during decision-making. It is drawn from the ensemble classifiers using the techniques of bagging and boosting [48]. When the modal data sources are uncorrelated in terms of sampling rate, data dimensionality, this technique can be used. Even though there is no such proof that late fusion is better than early one, many researchers prefer late fusion over early fusion [57]. Optimal integration of modalities, involve rules such as the Bayes rule [114], max-fusion [115], average-fusion [116], a majority vote for this approach. Late fusion methods are common as they resemble human cognitive abilities and can be integrated to generate a single common decision. Once we go to the higher abstraction level, the importance of content decreases; hence, the actual fusion level plays a vital role [91]. A late fusion approach is schematically depicted in Fig 2.

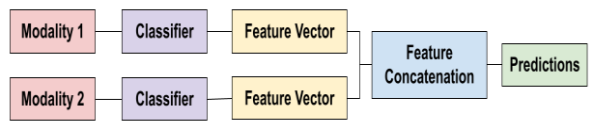

**FIGURE 2.** A late fusion approach

3) HYBRID FUSION

The deep neural network acts as a building block for intermediate fusion. It is the most widely used approach [117]. It changes input data to higher-level abstraction. Hybrid fusion learns a joint representation of different modalities. The fusion takes place at the commonly shared representation layer. The loss is propagated back to the feature extractor network during the training process [118]. Various modalities can be combined using slow or gradual fusion. However, this kind of fusion may lead to model overfitting and fail to learn the different correlations. Deep multimodal fusion performance is improved by reducing the data dimensions. After constructing a shared representation layer, PCA [107] and auto encoders [52] are used. Hybrid data fusion is far superior to early and late fusion. A "slow or gradual fusion" approach to integrating multiple fusion layers by fusing modalities from their high to low contribution performs well [119]. A hybrid fusion approach is depicted in Fig 3. A comparison of early, late, and hybrid fusion approaches is shown in Table IV.

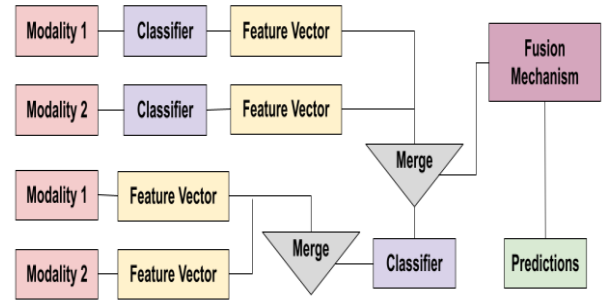

**FIGURE 3.** A hybrid fusion approach

TABLE IV
COMPARISION BETWEEN EARLY, LATE AND HYBRID FUSION TECHNIQUES BASED ON PROMINENT FEATURES

| Feature | Early | Late | Hybrid |
| --- | --- | --- | --- |
| Handling missing modalities | No | Yes | No |
| Inter features | Yes | No | Yes |
| Rich features | No | No | Yes |
| Labeled data | Yes | No | Yes |
| Multiple training | No | Yes | No |
| Design | No | No | Yes |
| Flexibility of fusion | No | No | Yes |

C. RECENT DATA FUSION TECHNIQUES

In the last few years, more sophisticated approaches to fuse multimodal data are reported in the literature. These are presented in this section. The fusion of unimodal embedding spaces into a common joint or shared representation possessing knowledge of semantic visual attributes and contextual language features is necessary to perform various multimodal integrated tasks [120]. The most common and widely used multimodal fusion pooling techniques reported in the literature are concatenation, element-wise multiplication, and weighted sum [10]. The model learns intra model features than the intermodal one; the greedy layer-wise pretraining approach is also used in different settings [52]. The tensor fusion network, where the unimodal, bimodal, and trimodal interactions are modeled using a 3-fold Cartesian product, is presented in [121]. It imposes high computational requirements and complexity on the system. All the modalities are used without any extraction. The low-rank multimodal fusion technique [122] addresses the shortfall of tensor fusion networks by using low-rank tensors for fusion but results in a complex architecture with a lot of computation and processing. Tensor-based multimodal fusion techniques provide excellent performance, but some approaches only consider "bilinear or trilinear pooling," which considers high-order correlations but results in very high dimensionality issues large outer product computation. [123], [124], which lacks exploiting the multilinear fusion power [125]. Feature fusion at a single instance ignores the local inter-model correlations, leading to performance degradation [125]. A slow, gradual fusion approach fusing modalities from higher to lower predictive power is recommended [119]. Multimodal deep learning performance is improved by maximizing the variation in mutual information of different channels [126]. Other approaches such as Multimodal Tucker fusion decompose and form a low-rank matrix decomposition overcomes bilinear polling's drawbacks by reducing complexity through tucker decomposition [127]. Memory-based fusion for multi-view sequential learning models the modality-specific and cross-modal interactions for multi-view datasets in [128]. Dynamic adaptive fusion scheme where the network decides the optimal way to fuse the modalities dynamically is presented in [129]. Cross-modal fusion by exploiting correlation across modalities by exchanging modality sub-networks is interpretable to a large extent [130]. Channel exchanging fusion that can model the inter and intra model tradeoff during fusion with a parameter-free dynamic approach and sub-network channel exchange [131] are proposed. The multimodal fusion task has also been modeled as a neural architecture search algorithm to find an appropriate search space and a suitable architecture to fuse the modalities [132]. A neural network-based model architecture based on global workspace theory from cognitive science is proposed to cope with uncertainties in data fusion with attention models across different modalities [133]. Deep-HOSeq Deep network with higher-order common and unique sequence information is proposed for sentiment analysis that models the inter and intra modalities with no reliance on attention mechanism [134]. Deep learning-based multimodal data fusion results can be optimized by including the model explainability, interpretability, justification, and reasoning capabilities for better and improved performance and predictions [135]. A summary of recent multimodal data fusion approaches is presented in Table V.

TABLE V
RECENT MULTIMODAL DATA FUSION TECHNIQUES

| Reference | Fusion Method | Specifications |
| --- | --- | --- |
| [131] | Channel exchange fusion | Models intra-model tradeoff with a dynamic approach |
| [129] | Dynamic fusion | An optimal dynamic adaptive fusion of modalities |
| [130] | Cross-modal fusion | Exploit cross-modal correlations through channel exchange |
| [136] | Architectural search | Optimal search space to fuse different modalities |
| [122] | Low-rank multimodal fusion | Use low-rank tensors to overcome the dimension and complexity of tensor-based fusion |
| [128] | Memory-based fusion | Model both modality-specific and cross-modal interactions |
| [121] | Tensor fusion network | Model inter and intra model dynamics |
| [123], [125], [127] | Bilinear Pooling Tucker fusion Multilinear fusion | Bilinear and trilinear tensor-based approaches by exploiting multilinear features |

This section provided a review of the traditional and recent methods for multi-modal data fusion techniques. Various multimodal applications are also provided. However, all the ways, especially the AI-based approaches, employ the black-box models, which are hard to interpret. The underlying functioning of these networks is not evident, and it becomes difficult to justify the outcomes of such models. It is necessary to bring in the explainable AI techniques to understand and explain such methods working and processes. It is even more challenging in a multimodal setting due to various scales of representations, alignment and resolutions than a single modality setting, as discussed previously. In the next section, we present the review of the existing explainable AI methods and their applications in multimodal AI tasks, with specific reference to vision and language tasks.

### III. EXPLAINABLE AI (XAI)

XAI is a multidisciplinary field involving different perspectives from social science, cognitive science, psychology, and human-computer interaction[56]. Artificial Intelligence systems powered by deep neural nets have achieved state-of-the-art results in various domains including, Computer Vision, Natural Language Processing, and Speech recognition [137]. But the primary focus lies on building intelligent systems achieving higher accuracy and predictive power, neglecting the trust and transparency aspect [138]. The underlying complexity and hidden layer processing in the deep neural nets make them opaque and black box models with an accuracy and interpretability tradeoff, i.e., more performing models are less interpretable [73]. It is hard to understand, interpret, and explain these models' internal processing and decision-making processes, limiting their social acceptance and usability [27]. In general, systems are interpretable if humans understand and interpret their working mechanism and decision-making process by asking questions like why the system made a particular prediction? Why answer the interpretability aspect, and how justifies how the system came up to a specific decision answer the explainability part [25].

"Interpretability is the degree to which a human can understand the cause of a decision and can consistently predict the model's results" [6]. Deep neural nets lack interpretability as it is difficult to analyze which modalities or features are driving the predictions [26].In real-world scenarios impacting human lives by automated algorithmic decision assistance such as in legal, healthcare, finance, transport, military, and autonomous vehicles, we expect AI systems to provide their predictions with proper evidence and justification [29]. The system's explanation should be human interpretable and understandable, mapping the human mental model to build trust, transparency, reliability for success and failure, robust, fair, and unbiased applications underlying ethical machine learning principles [139], [140]. Explainability is a legal concern to comply with the EU General Data Protection and Regulation (GDPR) act asking for" Right to explanation" to the users of an automated decision-making system [141]. To understand the inner working and learning mechanism, model debugging- to analyze the right/wrong predictions, design improvement, detect and mitigate adversarial attacks [16], i.e., artifacts in the datasets required in handling bias and fairness issues in the models explainability has become a prime concern to be addressed. The XAI field is continuously evolving and is critical in developing new AI algorithms and methods to explore how the models work and why they succeed or fail, to improve their design driving us towards the responsible AI paradigm of the future [24].

AI models are often right but sometimes for the wrong reason. However, the justification for the decision is often absent or vague [142]. Explainability in deep neural nets can be introduced in three different model training and development stages, namely in pre modeling, modeling, and post-modeling phase [24].

## A. LEVELS OF EXPLANATION MODELLING

1) Pre modeling-The explainability is included before the model development process. It primarily involves understanding and describing the data using exploratory data analysis, dataset documentation, dataset summarization, explainable feature engineering techniques, data collection without biases, and good experimental design to ensure clarity.
2) During modelling-Development of inherently explainable models, such models are explainable by design and typically employ intrinsic methods. Hybrid models such as deep KNN, Contextual Explanation Network [143], Self-Explaining Neural Network [144], joint prediction and explanation, and architectural adjustments through regularization techniques [145] are examples.
3) Post hoc Modelling is implemented after the model is developed by extracting explanations for already developed models through perturbations [146], Backpropagation methods [14], and proxy models [147], as it can be freely applied to any model without any constraints.

## B. SCOPE OF EXPLANATION

The explanation's scope is either local or global, depending on whether the explanation is derived from a single data instance or the entire model. Often local explanations are preferred over the global as they derive predictions for a particular data point rather than the model as a whole. XAI taxonomy is represented in Fig4. depicting the explainability stages, scope, and working principle [25].

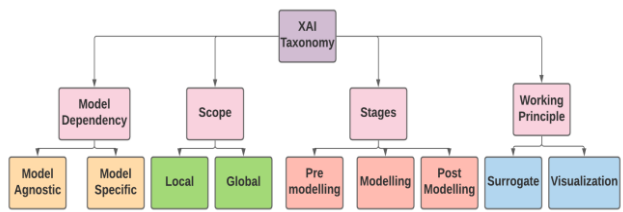

FIGURE 4. XAI Taxonomy

1) MODEL SPECIFIC EXPLANATION
Model-specific explanations apply to a specific model. Human interpretable explanations such as gradient-based methods [148], [149]have been proposed for the convolution neural networks. However, they do not focus on the entire region in an image to address a query.

2) MODEL AGNOSTIC EXPLANATION
Model agnostic methods [150], [151] that are independent and irrespective of the model have general modularity in design and can be applied for other domains. Various post hoc explainability approaches, such as partial dependent plot (PDP) [152], show the marginal effect of the final predictions' features. Individual conditional expectation (ICE) plot [153] visualize the dependence of features on the final prediction for unique data points, providing more insights than the PDP approach of averaging on all data points. Permutation feature importance works because a particular feature is important if alteration of that feature results in large model errors. Local interpretable model-agnostic explanations (LIME) is a perturbation-based method applied for a single data point and observes a change in the output based on the corresponding shift in input [151]. Shapley values [154] are used when each data point's contribution is variable in the final prediction. However, they work in a collaborative environment for deriving final predictions. Prototypes and criticisms help in minimizing the overgeneralization of the dataset. A prototype is a data point that represents the entire dataset. Criticism is a data point that is not well represented by the prototype; both prototype and criticism describe the data and lead to interpretable predictions [155]. Influence functions show the influence of each feature on the final prediction. They are used to understand model behavior, model debugging, detect dataset errors, and even create adversarial attacks [156]. In [157], model agnostic D-RISE, a visual explanation method for the predictions of object detectors irrespective of the model's inner working is presented. Human importance aware network-tuning (HINT) [158] method improves visual grounding by attending to the same visual features in an image which humans find important for predictions are few further examples of model agnostic methods.

## C. FEATURE ATTRIBUTION BASED METHOD

Feature attribution-based methods highlight image regions that are significant contributors in decision-making; however, it lacks semantic reasoning and interactions. Following methods are reported in the literature.

*Visualization techniques* focus on highlighting the input features most contributing and affecting the model's output. They are classified into back propagation-based and perturbation-based methods.

*Back propagation-based methods* look for relevant features based on gradients passed through the network. Visualization techniques such as weighted activations in Class Activation Mapping (CAM).

*Gradient-based methods* like saliency maps that focus on pixel intensities are based on high contributing features [148]. Gradient Input [159] improves the sharpness of the attribution maps. The attribution is computed using the input's partial derivatives concerning the input and multiplying them by themselves. Grad-CAM [14] captures class-specific gradient information in layers to produce a localization map of important features. Integrated Gradient [160] computes the model's prediction output gradient to its input features and requires no modification to the original deep neural network. DeepLift [159] calculates contribution scores based on reference activations or methods based on mathematical decomposition. Layer wise-Relevance Propagation (LRP) [161] computes backward relevance propagation to highlight the contributing features. Shapley Additive explanations (SHAP) [154]. Shap is the average contribution of all data points in a prediction. All these methods require access to the model parameters and thus an understanding of the model architecture.

*Perturbations-based Methods:* Visualize feature relevance by comparing output between inputs and altered or changed copy of the input like Occlusion sensitivity, RISE [150], LIME [144], etc. to identify sensitive features for prediction are proposed.

### D. DISTILLATION METHODS

Distillation methods are classified into local approximation models that build an approximate local model to derive insights for predictions based on a single data point example LIME. The model translation method builds a surrogate model on top of the original model whose interpretation is expected. E.g., decision trees that are inherently explainable.

### E. INTRINSIC METHOD

Intrinsic methods are inherently explainable. They self-explain using models' attention mechanisms that focus on important visual and textual regions. Joint training approaches that jointly model the predictions and explanations fall under this category and are further discussed in section 5. Fig 5. represents the taxonomy of various unimodal and multimodal deep explainability methods, which are presented here. Different deep learning-based XAI techniques are presented in Table VI.

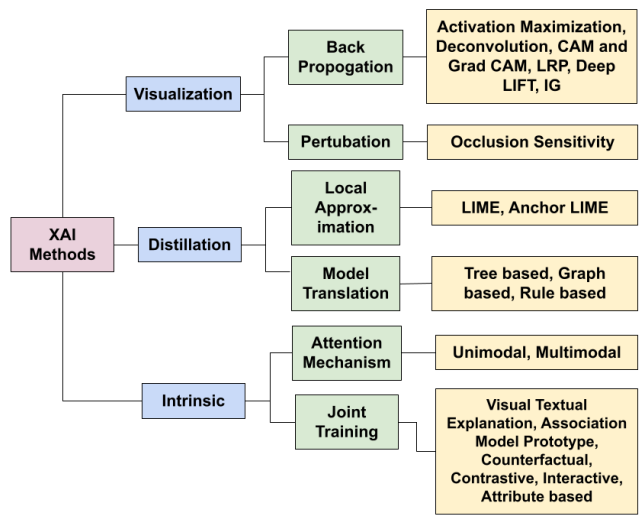

**FIGURE 5.** Methods for explaining deep neural nets

TABLE VI
DIFFERENT XAI TECHNIQUES FOR EXPLAINING DEEP NEURAL NETS

| Reference | XAI Method | Specification |
|---|---|---|
| | Backpropagation methods | Feature relevance by gradients passed through the network. |
| [162] | Activation maximization | Activation maximization activates few neurons over the others that are vital for making predictions. |
| [14] | Deconvolution | improves image estimation intensity through reversing convolution. |
| [130] | Grad CAM | captures class-specific gradient information in layers to produce a localization map of important features. |
| [131] | Layer wise Relevance Propagation | computes backward relevance propagation to highlight the contributing features. |
| [135] | Deep Lift | calculates contribution scores based on reference activations. |
| [135] | Integrated Gradients | adds the model's prediction output gradient to its input features and requires no modification to the original deep neural network. |
| [163] | Occlusion sensitivity | compares the output with input and altered copy of input data to identify sensitive features for prediction. |
| [151], [154], [158] | Local approximation LIME ,SHAP,HINT | Lime models change in prediction by a change in input for a local data point; Shap is the average contribution of all data points in a prediction. HINT looks at the same image regions as humans to make predictions. |

| [54], [147], [164] | Model Translation Tree, Graph, Rule-based methods | Builds a proxy model on top of the model whose explanations are desired, like a self-explanatory decision tree. |
|---|---|---|
| [165] | Attention Mechanism | Focuses and attends the regions that are important for making predictions. |
| [4],[63],[166] | Joint training | Jointly trains the explanation task with a prediction task to build the explainable model. |

## IV. EXPLAINABILITY IN MULTIMODAL DATA

### A. SIGNIFICANCE OF EXPLAINABILITY IN MULTIMODAL DATA

Multimodal explanations play a vital role in building intelligent systems powered with understanding and reasoning capabilities inherent and integral to humans [167]. In real-world settings, systems that are performing and explainable are desired. Unimodal vision or language systems offer either image-based visualizations of important input features or text-based post hoc justifications incapable of providing introspection and reason in multiple situations and scenarios [168]. In contrast, the multimodal setting of explanation explores the complementary and explanatory strengths in the different modalities, leveraging improved explanations that can justify, localize the evidence better supporting the final decision and offer significant benefits over unimodal approaches [4]. In specific scenarios, language modality may provide more insights and valuable information than the visual one to understand the concept and rationale better and vice versa [4], showcasing the complementary behavior of modalities in which one modality assist in enhancing the performance of the other. Multimodal sources extract more and comprehensive information from varied sources. Hence, they can offer diagnostic strengths that help understand the model working mechanism, model debugging i.e.to identify flaws in the model and ensure whether the model works as intended. The multimodal explainability models can also identify adversarial attacks and defense mechanisms [169], fairness and bias [13], providing scope for troubleshooting, rectification, improving overall model performance, predictive and explanatory power.

The primary explainable goal for an opaque, black box explanatory system such as deep neural nets is to answer how and why the model makes a certain prediction by inspecting the driving factors behind their decisions [170]. In reality, it is found that many explanatory systems are based on scientific modeling than explanation generation [171] and hence are not user-centric and role-based. They do not satisfy the needs of different stakeholders [139]. For instance, consider the task of interpreting an X-ray for diagnostic purposes; the radiologists are keen on mapping the visual evidence in the image to predict the diagnosis. On the other hand, the general clinician is more interested in the final text-based justification of the case. This demands justifications that are role-based, satisfying the requirements of different stakeholders simultaneously, and show the interplay between multimodal interactions required for building interactive explanation interfaces followed with feedback mechanism able to incorporate change in the system [96]. Evolution of vision and language tasks transformed from simple tasks requiring processing of fused multimodal embeddings such as image captioning. VQA to complex tasks such as visual common sense reasoning requires higher-order reasoning and a deep understanding of semantic context [172]. Such tasks demand the model to comprehend natural language and identify objects in the scene and capture inherent relationships between individual entities present in the input. The model's ability to reason their predictions has become essential, leading to the emergence of Explainable AI for multimodal settings [63]. Human visual explanations can help systems know where to attend human textual explanations and how they attended image regions to complement multimodal explanations[7].

### B. INTROSPECTION AND JUSTIFICATION SYSTEMS

Deep learning-based explanatory systems are broadly classified into justification [17] or introspection-based systems [163], [173]. Humans understand the justification-based explanations, but it lacks in deriving the causal aspect and interpretation [32]. The introspection-based systems focus on the network's internal behavior but are not well understood by humans. In real-world settings, justification, and introspection, both systems are desired [168] as the generation of lucid explanations improves human understanding and performance significantly [174].

## V. MULTIMODAL EXPLAINATION METHODS

In this section, we provide the classification of deep learning-based multimodal explanation techniques based on the approach they follow to explain visual and textual modalities into attention-based, counterfactuality-based methods, interactive approaches, and attribute-based methods following the taxonomy represented in Fig. 5.

### A. ATTENTION BASED APPROACHES

Attention-based approaches focus on certain factors in the data more than others by assigning more weight and importance. In multimodal tasks, such as visual captioning, visual question answering, or visual entailment, attention mechanisms play a crucial role in the alignment and fusion across different modalities [93]. In the attention-based approaches, such as in visualization methods like Grad-CAM, the attention features provide the explanation. Attention mechanisms are primarily used for the VQA task to attend or focus the image region on answering the query correctly. The VQA task of answering free-form natural language questions about images is explored in [175]. Efforts primarily focus on building interpretable models with a specific interest in exploring the input images or text in questions. The VQA model looks at these texts or images while answering the question. In [176], guided backpropagation and occlusion-based visualization techniques are employed to showcase vital regions on the image and text on which the VQA model focuses. To validate if explanations make VQA models more interpretable to humans in [15], the authors have proposed metrics for failure and knowledge prediction and found that human in loop (HIL) approaches make the model more understandable to humans.

The pioneering work in generating multimodal explanations was proposed by Park et al. [4] based on Teaching AI to Explain its Decision (TED) approach [166]. "The Pointing and Justification model" (PJ-X) for the visual question answering using VQA-X and activity recognition task using ACT-X datasets use attention mechanisms to explain the answer of a VQA task with textual explanations and corresponding visual regions. They pass attention masks between modules and explore modalities' complementary and diagnostic strengths, emphasizing the value-generating multimodal explanations. But this kind of model provides an indirect, inconsistent explanation of the model's internal working; identifying model flaws and requires a dataset that is augmented with explanations annotations. In [177], authors developed a multimodal approach generating explanations supporting deep network decisions in attentive pointing maps and text. In this model, a clinical diagnostic decision is conveyed with visual pointing and textual explanation in a coordinated fashion with a "visual word constraint" model. In [178], authors proposed a faithful multimodal explanation framework with a "bottom-up and top-down attention mechanism" to have consistent visual and textual explanations by segmenting the image for precise localization, demonstrating rationales improve human understanding and quality of explanations. In [179], the authors used a supervised attention model that trains human rationales to generate explanations. The VQA task requires different capabilities at varying levels. [180] discusses the state-of-the-art VQA models perform well on perception and reasoning questions but provide inconsistent explanations modeling a task. "Sub-Question Importance-aware Network Tuning (SQuINT)" forces the model to attend the similar areas of the image when answering the reasoning and the associated perception sub-question, thereby improving performance and consistency. Efforts have been made to leverage explainability in the task of VQA by using attention-based textual and visual explanations using parse trees. Hierarchical patterns to provide valid explanations and answer-specific sub-streams in sequential data using visual-textual attention.

In [181], a module for spatial grounding in VQA is proposed. This model addresses several visual recognition challenges, including the ability to infer human intent, the reason both locally and globally about the image, and effectively combines visual, language, and spatial inputs. In [182], authors evaluated the effect of explanations on the user's mental model for the VQA task by proposing an explainable VQA system using spatial and object features using the BERT language model on user perception of competency. They generated visual and textual explanations to complement the knowledge base, enhancing the model's prediction and interpretability. Multiple tasks, image attention, and knowledge base improve the overall model performance. In [183], authors have introduced a "self-critical training" method that ensures that image explanations of correct answers map more than other competitors' image areas mapping with the human mental model and decrease the incorrect answer probability in this region. In [184], the joint probability in VQA is maximized by "Hierarchical Feature Network (HF-Net)," where each hierarchical feature combines the attention maps with low-level semantics. Textual explanations are also generated for self-driving cars using an attention mechanism [79] and video description tasks. In [185], the proposed action justification model is based on the common-sense evidence using conditional variation autoencoder (CVAE), which provides better results than attention approaches and has improved grounding between humans and agents. In [186], authors have proposed an approach to enhance VQA performance by comparing competing explanations.

In [22], Zellers et al. has provided explanations for visual common sense reasoning through multiple choices. They proposed a Visual Commonsense Reasoning (VCR) task that answers a text question based on an image and provides reasoning accordingly. Both the answers and justifications are provided in multi-way generating explanations along with decisions. Thus VCR is more suitable to be applied for prototype model debugging to audit the model reasoning process. In [187], the multiword answer and rationale model for ViQAR are generated that goes beyond VQA in abstraction and reasoning abilities. An interpretable prediction attention-based mechanism is recently used in [161] for predicting anticancer compound sensitivity using attention-based convolution encoders for the task of drug discovery.

Despite the wide use and benefits exploration, challenges persist with attention mechanism as visual explanations generated using attention mechanisms do not explain if the model attends the right area. The region in the image to be focused on answering a particular question is not fixed. Lack of ground truth for evaluation of explanations imposes several restrictions. Attention maps do not look for the same area as humans do. Attention's explanatory power is questionable [188] as they lack an association with the attention weights and gradients mapping for generating faithful explanations.

### B. COUNTERFACTUAL EXPLANATIONS

Human thinking is contrastive and causal in the form of cause and effect. We ask why a particular X decision is taken and why not Y instead. For example, suppose a specific loan application is rejected. In that case, we are more interested in the measures and minimum possible changes to be undertaken to flip the decision to be accepted in the future. Multimodal explanations based on counterfactuality provide recommendations that provide actionable insights and recourse [189]. Specifying the minimal desired changes required to flip the decision in favor of the user, mapping well with the human mental model leveraging the class-specific and discriminative features and enhancing the model trust, transparency, accountability, reliability, social acceptance, and usability [190]. Interactive machine learning with a human-centered AI approach paves the way towards multimodal causal learning[32]. Some efforts in this direction are highlighted. In a vision-language setting, the visual explanation is the region with high positiveness or negativeness to measure how the target classifier changes corresponding to the negative class when a specific area is removed from the input using accuracy. The textual explanation is compatible with the visual counterpart and measures how the target classifier changes corresponding to the negative class when a specific region is removed from the input using accuracy [191].

In [18], a textual explanation model is proposed to investigate why the model predicted a class x instead of class X based on counterfactuals. They offered a "phrase-critic model" using explanation annotations and its counterparts. The model improves the textual explanation quality, but explanations are not accurate and faithful to the underlying model. In [192], a counterfactual visual explanation is generated based on the paper by Hendricks et al.[18]. The explanations are directly generated from the base model from the model's neurons and are accurate for the underlying model, and are without additional attribute annotations. In [168], a multimodal data classification model is built to classify a video to a particular class and justify why it is not classified to the other class-based on counterfactuality. Kanehira et al [191] developed a complementary explanation model by applying a joint training approach to generate a prediction and explanation, maximize the modality interaction information, and ensure that the explanations are complementary to each other. There is persistent language bias in VQA models. In [193], a counterfactual setting is employed where visual ground truth input is considered to be absent in a particular case. Similarly, studying the effect of visual biases synthesized similar but different images than ground truth that learned how and why the output change with visual distortions. For visual captioning, counterfactual explanations help analyze the models' working mechanism and the reasons behind certain predictions. Counterfactual explanations emphasize that observations, present or missing, lead to a specific output. In [194], counterfactual resilience in image descriptions is obtained by parsing entities, semantic attributes, and color information separately. Contrastive learning provides neural models with self-supervised competence using relevant and irrelevant pairs. It improves multimodal representations in pretraining handling noise and bias in the data[195].

### C. INTERACTIVE APPROACHES

An interactive approach to explanation leverages transparency in machine learning systems [196]. Interpretability is also explored by analyzing the accuracy in prediction for VQA in interactive systems. In [197], an interactive active attention-based model is proposed, that alters the model attention and provides user feedback if the forecast is incorrect by combining model explanation and annotation and evaluating the model explanation on the metric of user trust, mental model, and usability. In [96], the authors suggested the use of virtual agents to generate better multimodal explanations. Alipour et al. [197], evaluated multimodal explanations for VQA with and without explanations and established that explanations improve accuracy if the VQA system is wrong. They introduced an "active attention" mechanism, considering different attention maps. The generated explanations and annotations with a feedback loop are combined to rectify wrong answers [197]. In [198], the authors have proposed XAlgo, an interactive algorithm explaining the system's internal state through question answering. Explanation-based conversational systems are designed to provide better explanations than conventional report-based systems for customer relationship management with an interactive approach for multisensory fusion [199].

### D. GRAPH BASED APPROACHES

Reasoning out the question-answering task is carried out with learning question-specific graph-based interactions in the image scene graph [200]. Visual Reasoning task requires machines to ideally look beyond the face value of any image to capture correct relations and context before generating suitable descriptions. In [201], semantic attributes are present in the scene with semantic bottleneck based on context. The Context Semantically Interpretable Bottleneck (CSIB) provides a clear and interpretable explanation of each prediction by making the decision process of CNN more interpretable.

1) EXPLAINATION USING SCENE GRAPHS
The scene graph for an image is the graphical representation of its contents. The nodes are the depicted objects, and the edges are the relationships between them. In [202], graphs with only objects and relations are generated. [175] uses a multimodal approach for generating textual explanations for the visual question answering task using both image and language modalities, without collecting any additional data, and generated natural language explanation for VQA using scene graph and visual attention mechanism. With two variants based on region descriptors, objects, and relations, proved multimodal efficacy approach empirically. This approach doesn't rely on manually generated explanation data, as they use already available annotations from scene graphs addressing the PJ-X model's drawbacks, proposed by Park et al. [4].

2) KNOWLEDGE GRAPHS
Knowledge graphs significantly increase interpretability and explainability through semantic information and domain knowledge base infusion in healthcare and educational sectors [203]. A "Multimodal Knowledge-aware Hierarchical Attention Network "in which a knowledge graph with multiple modalities and different features is built for the medical field. In [204], a comprehensive view on the neuro symbolic AI perceptive is provided and integration of knowledge graphs in deep learning models for model interpretability is proposed.

## E. ATTRIBUTE BASED METHODS

The significance of attributes in providing explanations is far more critical, and there are efforts in this direction. In [205], the authors proposed an explanation model using attributes, counter attributes with examples and counter visual examples. To establish the intuitiveness of attributes in building class discriminative attribute-based multimodal explanations associating visual features with attribute information and better text image grounding for the visual common-sense reasoning task. In [206], the authors proposed the multimodal approach for generating the textual justification from the visual image attributes to represent and explain novel concepts with minimal supervision. Robust representations are generalized across different tasks using a spatiotemporal attention mechanism and a joint training approach for visual and textual modality in the zero-shot learning paradigm. In [207], the FLEX model is introduced that generates faithful language explanation for CNN decisions even for unseen class and provides a rationale. [201] proposed "Semantically Interpretable Activation Maps (SIAM)."

Attribute maps are linearly combined to see what different features the model has learned and how particular regions enhance interpretability. Visual common sense reasoning uses other object detection techniques for better text to image grounding and assigns attributes to object grounding with fewer parameters. Activation maps are used in [208] by considering different class pairs as complementary. Therefore, they can provide more discriminative cues to generate CAM using representative classes with discriminative cues using complementary regions activated for better and accurate CAM generation. In [209], a model produces attribute-based textual and visual explanations improving user trust and model flaws by generating complementary multimodal explanations. To justify a classification decision in zero-shot learning (ZSL) paradigm with a "joint visual attribute embedding and a multi-channel explanation module" to generate multimodal explanation is proposed. Different multimodal explanation generation approaches are represented in Table VI.

TABLE VI
MULTIMODAL EXPLAINATIONATION GENERATION APPROACHES

| Reference | Method | Specifications |
|---|---|---|
| [4],[184],[178], [210],[177], [181],[79],[176] ,[188],[211], [212],[19],[123] ,[213],[214], [215] | Attention-based approach | Explanations based on attention to capture the important visual and textual regions that contribute to a prediction. |
| [18],[192],[168] ,[191],[194], [190],[193],[32] | Counter Factual based | Explanations based on class-specific positiveness and negativeness of complementary modalities are counterfactuals. |
| [216],[203],[53] ,[204],[54] | Graph-based | Explanations are generated through knowledge graphs and scene graphs to make the models more interpretable. |
| [197],[182],[96] ,[196] | Interactive approach | Virtual agents' assistance, algorithms for an explanation of the systems' internal state, and feedback loop to rectify in case of a wrong prediction are widely used. |
| [217],[209] ,[218],[201], [205],[219], [220],[206]. | Attribute-based | Explanations provided with different attributes are intuitive due to their decomposability and interpretability and are explained with example-based explanations like humans. |

## VI. EXPLAINATION EVALUATION

Evaluation of explanations with metrics and methods is a major point of concern to leverage explainability and user confidence as they are task-specific and subjective [221]. Explanation evaluation tasks are categorized into the functionally grounded evaluation, such as depth of a decision tree evaluation. Human-grounded evaluation asks crowd workers for their preferences for better analysis and feedback. Application-grounded evaluation involves a human mental model – e.g., a human expert simultaneously explaining the outcome of the model [27]. Explanations are diverse and subjective, making their evaluation difficult without access to the ground truth. Evaluation of explanations without ground truth based on generalizing ability, persuasability is proposed in [222] with criteria for good explanation generalization performance, fidelity, faithfulness, relevance, per suability usability user satisfaction, and causality [223]. Human users can associate with the cause and effect scenarios. Better explanations involving novelty and completeness are desired[53]. Explanations shall support actual model behavior to be accurate. The model simulatability perceptive introduces a "leakage-adjusted simulatability (LAS)" metric for evaluating textual explanations making human users predict the output. Human interpretable evaluation of explanation is based on the accuracy, response time, consistency, and satisfaction established through simulation, verification, and counterfactuals [224].

Automated evaluation metrics such as BLEU-4 [225]is used for automatic evaluation of machine translation that is quick, inexpensive, and language-independent, that correlates highly with human evaluation, and that has a little marginal cost per run., METEOR [226] is an automatic machine translation metric used for unigram matching between the machine-produced translation and human-produced reference translations. ROUGE counts the number of overlapping units such as n-gram, word sequences, and word pairs between the computer-generated summary to be evaluated. The ideal summaries created by humans, CIDEr [227], are a human consensus-based image description evaluation metric for automatically evaluating image descriptions. SPICE [228] is a semantic propositional image caption evaluation metric for the automatic assessment of image captioning and improved evaluation compared to Cider, METEOR. Automated evaluation and human evaluation of explanation through crowdsourcing are followed for evaluating explanations in [178]. In [229], Explanations are evaluated based on the novel metric from the student-teacher paradigm that measures the degree to which the explanations help the student model simulate a teacher model in an improved way for better scalability for unseen data.

A schematic workflow pipeline of the fusion methods, tasks, applications, multimodal predictions, and explanation generation approaches is presented in Fig 6.

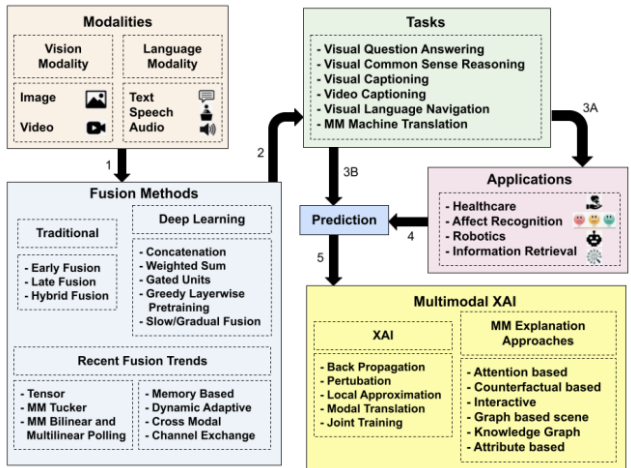

**FIGURE 6.** Multimodal explainability workflow pipeline

## VII. DIVERSE EXPLAINABILITY REQUIREMENTS

Explainability requirements are often user role and goal specific and hence are diverse [230]. For instance, model creators might require an understanding of how layers of a deep network respond to input data to debug or validate the model. In contrast, non-experts often need a functional explanation to understand how some output outside a model is produced. For instance, mode l examiners might require an understanding of how a model uses input data to predict to ensure that the model is trustworthy, not biased, or comply with the regulations [139]. Modular approaches have higher interpretability. The multimodal task of VQA becomes more interpretable and coherent with multimodal interpretability, wherein the answers can be justified in both textual and visual formats [210].

## VIII. DATASETS FOR VALIDATING THE EXPLAINABILITY IN MULTIMODAL NEURAL NETS

Multimodal XAI domain remained unexplored due to the lack of availability of task-specific datasets. Recently efforts in this direction are started. In [172], a new dataset is proposed based on Raven's Progressive Matrices (RPM) for the task of visual recognition reasoning, comprising images and related RPM problems, with tree-structured annotations. A counting-based dataset is sampled from the available VQA 2.0 and Visual Genome datasets for the task-specific release [200]. This work focused on countable quantitative question answering for answering specific queries asking how many? In [4], two novel datasets dedicated to explainability for visual question answering (VQA-X) and activity recognition (ACT-X) tasks comprising textual justifications for each image-text input pair. The VQA-X dataset has since then been considered a benchmark for many other explainable models. In [231], the VQA-e dataset is proposed for VQA predictions and explanations to improve overall performance. CUB dataset [232], an attribute-based zero-shot learning baseline dataset, is significant for attribute-based methods. VQA-CP [13] for multimodal bias evaluation are some popular multimodal explanation datasets. The dataset's availability makes the explainability goal more understandable and traceable in all contexts to leverage future research in the field.

## IX. MULTIMODAL BIAS AND FAIRNESS

Modalities exploit at different scales, and their contribution can vary, laying more importance on a certain modality over the other, providing suboptimal results. Imbalance data and feature selection introduce biases in models and machine learning algorithms, leading to a lack of fairness and transparency. Familiar sources of bias are through crowdsourcing workers and natural perceptions. Bias persists in word embeddings and at the sentence level. Algorithms often replicate and amplify the bias in the multimodal datasets. To detect and mitigate the bias in [233], proposed a regularization approach based on maximizing functional entropy and balancing modality contributions. Diversity in multimodal information often leads to such biases. In [234], how biases affect automated recruitment systems is demonstrated. Even after masking the inputs, gender and ethnicity discrimination based on a bias are present in the records. In the VQA task, the models often pick up statistical irregularities, introducing bias leading to memorizing rather than learning the task with a wrong evaluation [13]. Unimodal biases in the textual inputs neglect visual information impacting multimodal aspects. Such biases often lead to massive drops in performance when confronted with data outside training distributions [235]. Generalized and trivial questions are commonly answered with prior lingual knowledge instead of querying the image. Therefore, keyword dependencies over correct image reasons are necessary to obtain accurate, interpretable models and are comprehended via attention maps. For the task of image captioning, visual cues in the training images carry bias. Most models have a gender bias. Other such efforts focus on two significant subtasks or gender-neutral captioning in case of occlusions and correct gender classification otherwise. Such methods make multimodal frameworks more reliable, interpretable by allowing the models to provide reasonable predictions for the right reasons rather than looking for mere performance, looking at cause and effect aspects[193].

## X. ADVERSARIAL ATTACKS ENHANCE INTERPRETIBILITY

In addition to small changes or alterations, perturbations, called adversarial perturbations, result in adversarial examples leading to change in output. These adversarial examples can mislead the classifiers to make wrong classification decisions. They are so minute that human eyes often missed them. Adversarial examples can also be used for understanding neural networks. Masking visual modality to see the partial or complete influence of statistical language patterns through adversarial attacks classification by attributes, they are also natural candidates to study misclassification and robustness [205]. In the interpretation of adversarial examples, discriminative attributes predict the correct and wrong class predictions, and adversarial perturbation increases the network's robustness. In [236], attention-guided adversarial attacks for the VQA model are proposed. The show and fool algorithm for attacking visual grounding with adversarial examples shows that models can be fooled through adversarial attacks despite attention and localization, showcasing the need to establish strong defense and mitigation mechanisms against adversarial attacks[237].

## XI. OBSERVATIONS AND RECOMMENDATIONS

1) Existing unimodal vision or language systems offer either image-based visualizations of important input features or text-based post hoc justifications. Lack of complete introspection and justification highlights the need for further exploration of multimodal approaches for explainability.
2) Natural language explanations can be mutually inconsistent. There are persistent inconsistencies and biases in the multimodal data, such as in visual question answering. Measures to tackle prevalent bias and fairness issues are of utmost importance.
3) Even if the classifier is very accurate, without having access to complete explanations to understand how decisions are made in the model, it is not sure that it is making decisions for the right reasons. The model may learn things that should not affect generalization and focus on faithful explanations complete to model and substitute task.
4) Building trust by asking questions transparency and accountability is not enough; we need auditable, explainable, interactive, augmented, and learning from human feedback. Artificial intelligence supplemented with human intelligence with a human in the loop approaches for better evaluation should be proposed.
5) Confidence that human knows how and why the decision was made should follow with recourse in case of dissatisfaction. In case of disagreement with systems output, measures to change the same should be feasible, which can be accomplished with the interactive system design with a feedback loop.
6) Multimodal diagnostic and explanatory capabilities need to be further explored for inconsistencies to leverage comprehensive and fine-grained class-specific and discriminative feature understanding of relations and interpretation of explanations with hybrid, composite approaches to model multiple data modalities.
7) Multimodal explanation generated may not be correlated, complete and faithful, i.e., accurate to the model's internal decisions and processes. Hence, lack of proper grounding, reasoning, and context adaptation will focus on developing new models.
8) Explanations lack addressing the lay users, developers, domain experts, and stakeholders to simulate the human cognitive process and user-centric design. There is a need for interactive, user-friendly, trustworthy, and diverse multimodal explanations to satisfy different stakeholders' needs.
9) Explanations lack proper evaluations on bias, fairness, trustworthiness, fidelity, generalized ability, causality, completeness, novelty, and quality in mapping the human mental model due to the diversity aspect in the evaluation metrics task-specific subjective nature of explanation. Different automatic and human evaluation metrics with theoretical backgrounds need to be developed.
10) Despite the great insights into various explanation modes' efficacy, previous studies do not interactively involve the human subjects and feedback in producing these explanations. Involving human subjects can improve the quality and trust, and usability of explanation.
11) Multimodal explanations lack robustness and are prone to adversarial attacks through small input perturbations. An adversarial defense mechanism shall be established to combat the situation.
12) Measures to tackle the language bias and fairness issues in the data due to diverse information from multiple modalities are carried out on a broad scale.
13) There is a need to evaluate the multimodal explanations on the grounds of causality and visual grounding.
14) An explanation lack infusion of the prior domain knowledge base required in the decision-making process and explanation where knowledge graphs provide a promising direction.

## XII. DISCUSSIONS AND FUTURE DIRECTIONS

Multimodal research has achieved much success in various downstream tasks. However, there is still a long way to meet up to the human level performance as challenges persist with representation, alignment, translation, fusion, and co-learning. Lack of common sense and reasoning, contextual adaptation, labeled data requirements, development of novel and improved architectures, and evaluation metrics are still prevalent. No explanation theory is foundationally established and is more domain-specific, explaining with analogies and examples, multiple modalities, contrastive and counterfactuals explanations show promising edge in this way as an explanation is as vital as a prediction and are jointly modeled. The explanation structure shall be based on the human mental model and evaluated with a human in the loop approach to better understand outcomes. An enormous scope of improvement exists for building interpretable and explainable models for multimodal settings; instead of explaining the model predictions in a post hoc manner inherent and self-explaining model should be developed leveraging visual and textual modalities integrated benefits [238]. The type of explanation appealing to humans is not established; multimodal interactive explanations with user feedback can improve quality and user satisfaction. Interacting with explanations of machine learning models is an enabler for scientific discoveries for human-computer interaction. Due to the engagement of multiple modalities, multimodal explanations leverage the ability to satisfy different stakeholder's requirements with a global context enabling the development of vigilant AI systems that offer trust and improved transparency in the decision-making process. Combining knowledge-driven and data-driven approaches results in interpretability and accuracy in the models. Multimodal explanations involving different visual and textual modalities will better understand the semantic and context in a human-understandable format. Explainability in multimodal nets becomes a prime requirement in critical domains such as medical, legal, finance, autonomous vehicles, robotics, and other fields with diverse modality involvement. The user-centric interactive approach in design backed up with respective domain knowledge leads to quality explanations fostering new facets and a better understanding and traceability of model working and decisions with XAI in hand to achieve integrated interpretability.

Contrastive learning, probabilistic graphical models like causal networks and counterfactuals open a vast arena of interpretable and transparent deep learning algorithms, thereby reducing overall bias and increasing the system's reliability. Simultaneously, it opens the requirement for novel reasoning-based datasets for models and relevant metrics quantifying the separation from ground-truth data and measuring the higher-level reasoning and cognition capabilities over complex datasets. Adversarial attacks tend to enhance the system's robustness and interpretability, encountering the need for defense mechanisms.

Recently OPEN AI have reported multiple algorithms relevant to this body of work which are presented in next subsection.

OPEN AI DALL- E AND CLIP MODELS

Multimodal learning aims to learn concepts through several modalities. After GPT 3, open AI came up with a transformer-based architecture that combines image and language with Dall E and CLIP models. Unlike a biological neuron, a multimodal neuron can represent abstract features and concepts in a high-level theme compared to a single feature representation. Understanding the literal, symbolic and conceptual meaning is a crucial development in multimodal learning.

Dall E [239] is a transformer-based image generation model from text captions that fills the gap between vision and text for a wide range of concepts explained in natural language. It uses a 12-billion parameter version of the GPT-3 transformer model to interpret natural language inputs and generate corresponding images from text captions similar to CLIP building visual concepts through language. The second breakthrough in Multimodal AI introduces Open AI's CLIP model [232] trained on image sentence similarity scores connecting images and text model that stands for contrastive language image pretraining classifying image text pairs. Inspired by the zero-shot learning paradigm, it uses a single pretraining task to generalize to other interest domains. This multifaceted neuron's working is interpreted by different feature visualization and data example techniques with computational efficiency due to the transformer-based approach. Still, it cannot generalize for all the tasks. Despite the performance, the CLIP model is subject to associative bias in the underlying data. The Middle East neuron was associated with terrorism. An immigration neuron responded to Latin America targeting a specific group of people, and typographic adversarial attacks are reported on this model, raising severe concerns about its social acceptance. There is still an urgent requirement to analyze the models' societal impacts concerning the data biases that they carry. Multimodal explainability feature visualization and data examples techniques play a promising role in understanding the working mechanism and detecting and dealing with the underlying biases and adversarial attacks on these models.

OpenAI investigates their recent CLIP model's inner workings via faceted feature visualization and deduces findings that some neurons in the last layer respond to distinct concepts across multiple modalities. The neuron fire for photographs, drawings, and signs depicting the same concept, even when the images are vastly distinct. They identify and investigate neurons corresponding to persons, geographical regions, religions, emotions, and much more. Both DALL·E and CLIP represent significant advancements in transformer models. Indeed, they are important milestones for the computer vision community.

### XIII. CONCLUDING REMARKS

This review described and categorized the different explainability methods and techniques in a multimodal setting. The importance of explainability techniques concerning diverse image and text modalities in vision and language settings is fascinating. The context, reasoning, and semantic attributes harnessed by different explanation methods focus on the proper distinction. Crucial aspects of bias fairness and adversarial defense mechanisms aligned with explainability in multimodal nets are highlighted. To derive the full benefits of multimodal setting, new benchmarks and diagnostic datasets play a prime role. Explainability is a prominent aspect in a multimodal environment, establishing trust and transparency in the working mechanism understanding and tracing model flaws. Recently the pace with which the field is evolving, the upcoming developments are tremendous. We hope our survey is a step to provide a roadmap for further improvements and research directions in this active domain.

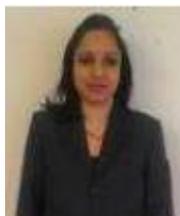
**GARGI JOSHI** received her Bachelor of Technology degree in computer science and engineering from Dr. BAMU, University, Aurangabad, in 2012 and Master of Engineering from University of Pune, in 2014 and currently a junior research fellow pursuing her PhD. in computer engineering, Artificial Intelligence, Deep Learning, Multimodal XAI domain from Symbiosis International Deemed University Pune. From 2012 to 2019, she worked as an Assistant Professor in D.Y. Patil College of Engineering, Ambi, Pune. She is interested and keen on recent advances in AI machine learning and deep learning technology.

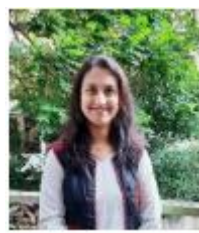
**RAHEE WALAMBE** received MPhil, Ph.D. Degree from Lancaster University, UK, in 2008. From 2008 to 2017, she was a Research Consultant with various organizations in the control and robotics domain. Since 2017, she has been working as an Associate Professor at Dept of Electronics and Telecommunications at Symbiosis Institute of Technology, Symbiosis International University, Pune, India. Her area of research is applied Deep Learning and AI in the field of Robotics and Healthcare.

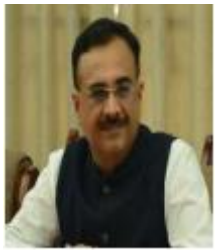

**KETAN KOTECHA** received his MTech and Ph.D. from the Indian Institute of Technology, Mumbai. He is the Head of Symbiosis Centre for Applied AI and Dean of Faulty of Engineering, Symbiosis International University, Pune, India. He is an expert in Artificial Intelligence and Machine Learning. He is the recipient of several awards and research grants in AI application areas.